\pdfoutput=1
\documentclass[11pt]{article}

\usepackage[preprint]{neurips_2023}

\usepackage{times}
\usepackage{latexsym}
\usepackage{makecell}
\usepackage[T1]{fontenc}
\usepackage[utf8]{inputenc}

\usepackage{microtype}
\usepackage{breakurl}
\usepackage{times}
\usepackage{latexsym}

\usepackage{algorithm}
\usepackage{algorithmic}

\usepackage{microtype}
\usepackage{hyperref}

\usepackage{graphicx}
\usepackage{xspace}
\usepackage{paralist}
\usepackage{booktabs}
\usepackage{multirow}
\usepackage{amsmath}
\usepackage{amsfonts}
\usepackage{amssymb}
\usepackage{pifont}
\usepackage{mathrsfs}
\usepackage{mdwlist}
\usepackage{enumitem}
\usepackage{color}
\usepackage{dsfont}
\usepackage{amsthm}
\usepackage{bm}
\usepackage{float}
\usepackage{caption}
\usepackage{subcaption}
\usepackage{hyperref}
\usepackage{tcolorbox}
\usepackage{tikz}
\newcommand*{\circled}[1]{\lower.7ex\hbox{\tikz\draw (0pt, 0pt)%
    circle (.5em) node {\makebox[1em][c]{\small #1}};}}

\newcommand{\notes}[1]{}%{\it {\small {#1}}}}

\theoremstyle{definition}

\theoremstyle{plain}

\newcommand{\ith}[1]{\ensuremath{i^{{th}}}}

\newcount\permx
\newcount\permy
\def\permdot#1#2{
\permx=#1 \advance\permx by-1
\permy=#2 \advance\permy by-1
\psframe[fillcolor=black, fillstyle=solid]
(\permx,\permy)(#1, #2)
}

\newcommand{\boxnum}[1]{{\setlength{\fboxsep}{1pt}\raisebox{1pt}{\hspace{1pt}\fbox{\tiny #1}\hspace{1pt}}}}
\newcommand{\ind}[1]{\ensuremath{_{\kern-0.5pt\boxnum{#1}}}}

\newcommand{\smallnt}[1]{\ensuremath{_{\mbox{\tiny PP}}}\xspace}

\newcommand{\pseudocode}{Algorithm}
\floatname{algorithm}{\pseudocode}

\iffalse

\else

\fi

\newcommand{\RNum}[1]{\uppercase\expandafter{\romannumeral #1\relax}}

\title{SpeechAlign: Aligning Speech Generation to \\Human Preferences}

\author{
    \textbf{Dong Zhang}\thanks{Equal contribution.} ,
    ~\textbf{Zhaowei Li}$^\ast$,
    ~\textbf{Shimin Li},
    ~\textbf{Xin Zhang},
    ~\textbf{Pengyu Wang},\\
    ~\textbf{Yaqian Zhou}\thanks{Corresponding author},
    ~\textbf{Xipeng Qiu}\footnotemark[\value{footnote}] \\
    Fudan University\\
    {\tt dongzhang22@m.fudan.edu.cn},
    {\tt lizhaowei126@gmail.com}\\
    {\tt 	\{zhouyaqian,xpqiu\}@fudan.edu.cn} \\
    \\\\
    \url{https://0nutation.github.io/SpeechAlign.github.io/}
}

\begin{document}

\maketitle

\begin{abstract}
Speech language models have significantly advanced in generating realistic speech, with neural codec language models standing out. However, the integration of human feedback to align speech outputs to human preferences is often neglected. This paper addresses this gap by first analyzing the distribution gap in codec language models, highlighting how it leads to discrepancies between the training and inference phases, which negatively affects performance.
Then we explore leveraging learning from human feedback to bridge the distribution gap.
We introduce SpeechAlign, an iterative self-improvement strategy that aligns speech language models to human preferences. 
SpeechAlign involves constructing a preference codec dataset contrasting golden codec tokens against synthetic tokens, followed by preference optimization to improve the codec language model. This cycle of improvement is carried out iteratively to steadily convert weak models to strong ones.
Through both subjective and objective evaluations, we show that SpeechAlign can bridge the distribution gap and facilitating continuous self-improvement of the speech language model. Moreover, SpeechAlign exhibits robust generalization capabilities and works for smaller models. Code and models will be available at ~\url{https://github.com/0nutation/SpeechGPT}.

% Speech language models has made great progress in speech generation. However, current methods ignore the integration of human feedback to align speech outputs to human preferences. 
% In this paper, we first analysis the distribution gap in codec language models and find that it can cause discrepancy between training and inference process and there adversely impacts the performance.
% To address these problems, we propose SpeechAlign, an iterative self-improving strategy that aligns speech language models to human preferences.
% Specifically, we first construct the codec preferences dataset by constracting the golden codec tokens and synthetic tokens. Then we conduct preference-aware optimization to align codec language model with human preferences. After that, we conduct this procedure iteratively and continuously improve the codec language model.
% Experiment results show that SpeechAlign can boost speech generation performance and make speech language model self-improve iteratively. SpeechAlign also shows strong generalization abilities and works for smaller models. Further analysis proves that SpeechAlign can bridge the modality gap.
\end{abstract}

\section{Introduction}
% Speech language models have achieved impressive results in produce high-quality speech in zero-shot manner~\citep{wang2023neural,borsos2022audiolm,yang2023uniaudio, le2023voicebox, zhang2024speechgptgen}, benefited from the availability of large amounts of speech data. 

\begin{figure*}[t] 
    \centering 
    \includegraphics[width=1\textwidth]{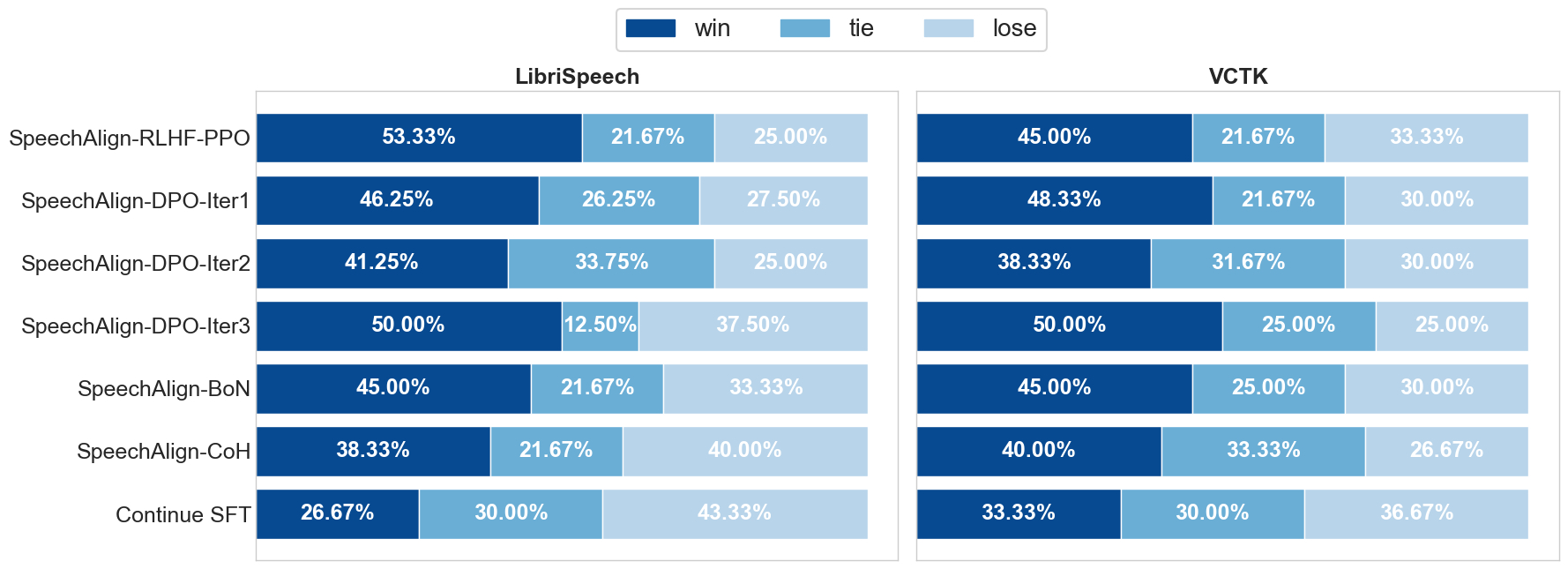} 
    \caption{Qualitative side-by-side comparsion results of preference optimized models versus the baseline SFT model on zero-shot text-to-speech performance. SpeechAlign-RLHF-PPO denotes models optimized by RLHF using PPO algorithm. SpeechAlign-DPO-Iter1 denotes models optimized by Direct Preference Optimization method at the first iteration. SpeechAlign-DPO-Iter2 and SpeechAlign-DPO-Iter3 denote the models optimized at the second and third iterations, respectively. SpeechAlign-CoH represents models optimized by Chain-of-Hindsight strategy. SpeechAlign-BoN refers to baseline SFT model employing Best-of-N sampling method.
    SpeechAlign-BoN, SpeechAlign-RLHF-PPO and SpeechAlign-DPO series models significantly outperform baseline model on both LibriSpeech and VCTK dataset.
    }
\label{fig:comparsion} 
% \vspace{-0.5em} 
\end{figure*}

Large language models (LLMs) have showcased their potent abilities through techniques such as pretraining, supervised fine-tuning (SFT), and Reinforcement Learning from Human Feedback (RLHF)~\citep{openai2023gpt4,touvron2023llama}. 
The field of speech language modeling has seen significant progress~\citep{wang2023neural,borsos2022audiolm,zhang-etal-2023-speechgpt}, particularly with the adoption of discrete speech representations~\citep{hsu2021hubert,zhang-etal-2023-dub} like audio codecs~\citep{défossez2022high,zeghidour2021soundstream, zhang2023speechtokenizer}.
% Similarly, many works have utilized discrete speech representations to build speech language models~\citep{wang2023neural,borsos2022audiolm,zhang-etal-2023-speechgpt}. Among these, neural codec language models based on audio codecs~\citep{défossez2022high,zeghidour2021soundstream, zhang2023speechtokenizer} have achieved impressive results in speech generation tasks~\citep{yang2023uniaudio,wang2023neural}. 
However, current speech language models primarily focus on the SFT stage associated with empowering the LLM's instruction-following capabilities, neglecting the integration of human feedback to align speech outputs to human preferences regarding quality, naturalness, and expressiveness.
Fortunately, \textit{learning from human feedback} has emerged as a powerful solution for aligning LLM output distribution with human expectation~\citep{stiennon2022learning,bai2022training,ouyang2022training}.
The most successful approach, reinforcement learning from human feedback (RLHF), achieves this by integrating rewarding modeling and a reinforcement learning phase. Additionally, some computationally efficient alternatives have proven to be effective in aligning LLM behavior without the need for explicit reward modeling~\citep{rafailov2023direct, zhang2023wisdom,wang2024inferaligner}. 

% 缺个逻辑承接句？
The key to the success of speech language models~\citep{wang2023neural,borsos2022audiolm,zhang-etal-2023-speechgpt} that build on LLMs is utilizing audio codecs that discretize the speech representations.
Neural Codec Language Models, leveraging audio codecs, have demonstrated their effectiveness in speech generation tasks~\citep{yang2023uniaudio,wang2023neural}.
It primarily utilizes a hierarchical approach that consists of a pipeline of autoregressive~(AR) and non-autoregressive~(NAR) models, as illustrated in Figure~\ref{fig:main} (a).
AR model generates semantic tokens~\citep{borsos2022audiolm} or the first layer of codec tokens~\citep{wang2023neural}, referred to as AR tokens. These AR tokens serve as input for NAR model to generate acoustic tokens~\citep{borsos2022audiolm} or subsequent layers of codec tokens~\citep{wang2023neural},  termed as NAR tokens. 
However, this pipeline system introduces a discrepancy between the training and inference phases for the codec language model. In training, NAR model is fed with \textbf{golden AR tokens} derived from real speech. However, the model receives \textbf{synthetic AR tokens} generated by the AR model during inference. As demonstrated in Section~\ref{preliminary}, there is a distribution gap between these two types of AR tokens, which adversely impacts the performance of the NAR model.

\textbf{Can we calibrate the output of codec language models to the authentic codec distribution by learning from human feedback}?
Collecting a large, high-quality preference dataset for codec language models is challenging. First, codec tokens are often represented in numerical form, which is not directly understandable by humans, making it impossible to collect human preferences for these tokens directly. Furthermore, collecting human preferences on speech to gather feedback on codec tokens poses multiple challenges, including inconsistency across various human annotators and the difficulty of scaling up the dataset size.
% Therefore, we adopt a self-improving strategy that aligns speech language models without the need for additional human-annotated data.

% We propose SpeechAlign, a speech language model aligned with human preferences. 
We propose SpeechAlign, an iterative self-improving strategy that aligns speech language models to human preferences.
To avoid the need for additional human-annotated data, we construct the pairwise preference codec dataset by considering golden AR tokens as preferred data and synthetic AR tokens as dis-preferred data. Human verification is conducted to ensure its consistency with human preferences. 
After obtaining the preference dataset, we explore different preference optimization strategies to improve codec language models. Following a complete cycle, we iteratively perform preference dataset collection and preference-aware optimization to convert weak codec language models to stronger ones continually.
Experimental results show that SpeechAlign can continually improve the speech generation performance of speech language models.

Our contributions are summarized below:
\begin{itemize}[itemsep=1pt, leftmargin=10pt, parsep=0pt, topsep=1pt]

    \item 
    We propose SpeechAlign, the first to align speech language models by learning from human feedback.

    \item 
    We propose an iterative self-improving strategy to convert weak codec language models to stronger ones without additional human-annotated data.

    \item 
    We analyze the issue of distribution gaps in codec language models and explore various strategies to bridge the gap.

\end{itemize}

\section{Preliminary Analysis on Distribution Gap}

In this section, we perform preliminary experiments to analysis the distribution gap between golden and synthetic codec tokens and demonstrate that this gap can degrade the performance of the codec language models.

\begin{figure*}[t] 
    \centering 
    \includegraphics[width=0.9\textwidth]{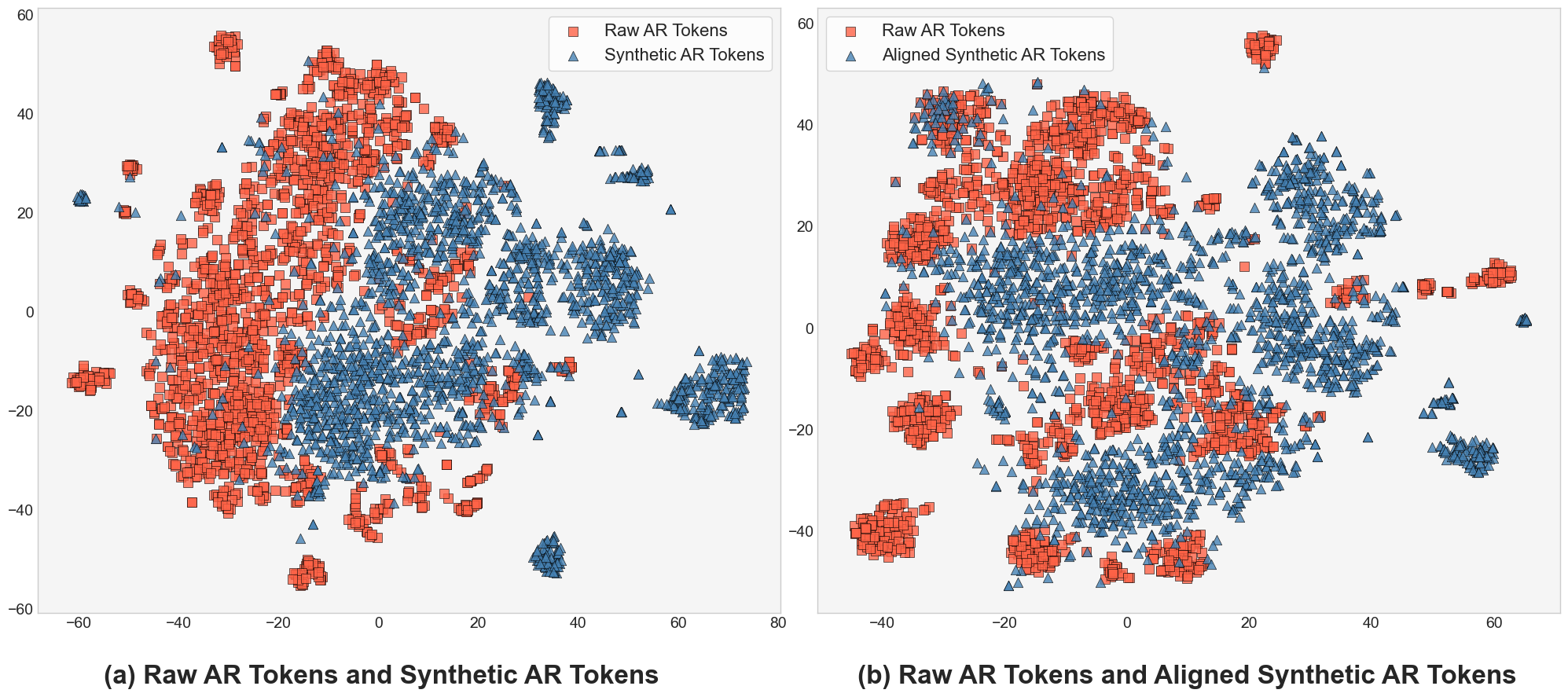} 
    \caption{T-SNE visualization of representations of different AR tokens. \textbf{Left}: Golden AR tokens and synthetic AR tokens. \textbf{Right}: Golden AR tokens and aligned synthetic AR tokens.}
\label{fig:tnse} 
% \vspace{-0.5em} 
\end{figure*}

\subsection{Background}~
\label{sec:pre:background}
We build a codec language model, referred to as \textit{SpeechAlign-sft}, serving as the baseline system to analysis the distribution gap.
Similar to~\citep{zhang2024speechgptgen,budzianowski2024pheme}, we rely on SpeechTokenizer~\citep{zhang2023speechtokenizer} to extract speech codec tokens.
SpeechTokenizer is a Residual Vector Quantization~(RVQ)-based speech tokenization method and hierarchically disentangles different aspects of speech information across different RVQ layers. The output of SpeechTokenizer comprises \(Q = 8\) hierarchical RVQ tokens \((q_1, \ldots, q_Q)\). 
\textit{SpeechAlign-sft} consists of a SpeechGPT~\citep{zhang-etal-2023-speechgpt}-based autoregressive~(AR) model and a SoundStorm~\citep{borsos2023soundstorm}-based non-autoregressive~(NAR) model.
The AR model learns the mapping from input golden text to the first layer of codec tokens $q_1$. We continue finetuning the pretrained SpeechGPT model in~\citep{zhan2024anygpt} on LibriSpeech dataset to get the AR model. Details about training process as described in Section~\ref{sec:041:setups}.
The NAR model adopts the training and inference procedure of SoundStorm~\citep{borsos2023soundstorm} and 
learns to generate subsequent layers of SpeechTokenizer tokens conditioning on the first layer tokens and prompt speech. We use the pretrained SoundStorm model in~\citep{zhan2024anygpt}. 
At inference time, the AR model converts input text to \textbf{AR tokens} and the NAR model uses these tokens along with prompt speech as conditions to generate \textbf{NAR tokens}. These tokens are then concatenated and converted into speech by the SpeechTokenizer decoder.

\subsection{Visualization of Distribution Gap}~
\label{sec:pre:visualize}
To analysis the distribution gap, we randomly select 1000 speech-text pairs from LibriSpeech dataset and construct a test corpus composed of triplets $ D_{vis} = \{(\mathbf{t},\mathbf{y_g},\mathbf{y_s})\}$ following the procedure in Section~\ref{sec:method:data}. Here $\mathbf{t}$ is the input text, $\mathbf{y_g}$ is the golden AR tokens and $\mathbf{y_s}$ is the synthetic AR tokens generated by SpeechAlign-sft.
The input text $\mathbf{t}$ is concatenated with the golden AR tokens $\mathbf{y_g}$ and fed into the SpeechAlign-sft model. This process yields the hidden states of the last layer for each AR token in the sequence. By applying mean pooling across the temporal dimension, these hidden states are aggregated to produce a single vector representation $Rep_g$ for the golden AR tokens.
Similarly, we acquire $Rep_s$ for the synthetic AR tokens using the same procedure.
The vectors are visualized in a 2D space using t-SNE, as shown in Figure~\ref{fig:tnse} (a). We can observe that the representations of golden AR tokens and synthetic AR tokens are so dissimilar that they naturally form two distinct clusters, indicating that significant distributional gap exists between them.

\subsection{Distribution Gap Degrades Performance}~
\label{sec:pre:downstream}
% The distribution gap between golden AR tokens and synthetic AR tokens causes discrepancy between training and inference process of NAR model, which may impact its performance. 
The NAR model is trained using golden AR tokens as input, but during inference, the input switches to synthetic AR tokens. This results in a discrepancy between the training and inference processes due to the existing distribution gap, potentially affecting performance.
To delve into this issue, we conduct a speech reconstruction experiment with NAR model.
We construct a dataset composed of triplet data $ D_{test} = \{(\mathbf{z},\mathbf{y_g},\mathbf{y_s})\}$, with $\mathbf{y_g}$ and $\mathbf{y_s}$ described in Section~\ref{sec:pre:visualize} and $\mathbf{z}$ represents 3-second prompt speech from the same speaker but distinct from the speech used for $\mathbf{y_g}$ and $\mathbf{y_s}$.
The NAR model performs speech reconstruction by taking prompt speech combined with either golden AR tokens or synthetic AR tokens as input, to generate speech for each type of tokens respectively.
The quality of the generated speech is evaluated based on the word error rate (WER) and speaker similarity (SIM) metrics, compared against the ground truth. As shown in Tabel~\ref{tab:preliminary}, speech generated from golden AR tokens exhibits superior WER and Speaker Similarity scores compared to that generated from synthetic AR tokens. This finding proves that the distribution gap adversely affects the NAR model's performance.

\begin{table}[t]
    \setlength{\abovecaptionskip}{0.3cm}
    \Large
    \centering
    \resizebox{0.4\columnwidth}{!}{
    \begin{tabular}{l|cc}
        \toprule
        Input & WER($\downarrow$) & SIM($\uparrow$) \\
        \midrule
        Groundtruth & 3.4 & -   \\
        \midrule
        Golden AR tokens & 5.9 & 0.93   \\
        Synthetic AR tokens & 7.2 & 0.87   \\
        \bottomrule
    \end{tabular}}
    \caption{Results of NAR model's speech reconstruction performance with different AR tokens as input.
    }
    \label{tab:preliminary}
    % \vspace{-0.3em}
\end{table}
\label{preliminary}

\section{SpeechAlign}

\begin{figure*}[t] 
    \centering 
    \includegraphics[width=1\textwidth]{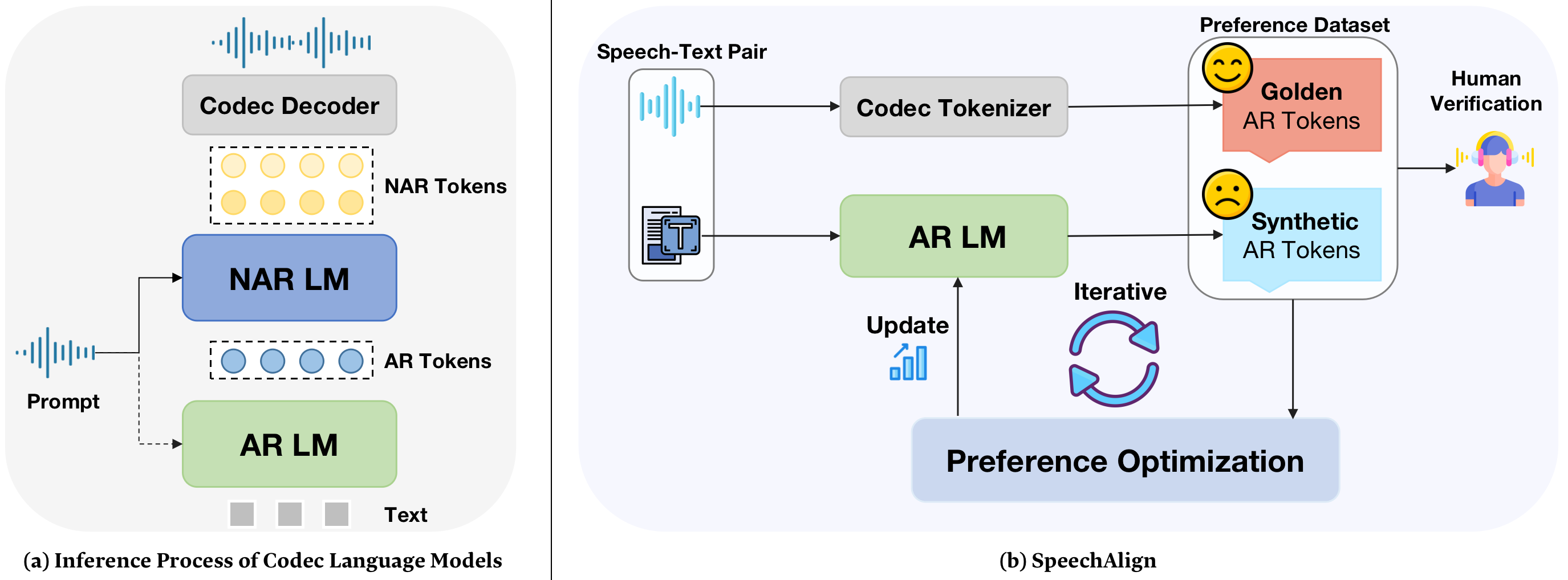} 
    \caption{AR LM refers to autoregressive models and NAR LM refers to non-autoregressive models. \textbf{Left}: Illustration of inference process of codec language models. \textbf{Right}:
    Illustration of SpeechAlign method.}
\label{fig:main} 
% \vspace{-0.5em} 
\end{figure*}

We take SpeechAlign-sft detailed in Section~\ref{sec:pre:background} as the baseline system, referred to as $p_{\theta_0}$. Within this framework, the AR model is represented as $p_{\theta_0}^{ar}$, and the NAR model as $p_{\theta_0}^{nar}$. 
As shown in Figure~\ref{fig:main} (b), the first step of SpeechAlign is to construct perference dataset that contrasts golden codec tokens with synthetic codec tokens. Utilizing this dataset, we implement various preference optimization strategies to align the baseline model. This process is iteratively executed, enabling the continuous self-improvement of codec language models.

\subsection{Preference Data Collection}~
\label{sec:method:data}
A standard method for collecting human preferences involves prompting the model to produce two distinct responses to a query, after which annotators are asked to select the one they prefer. However, collecting human preferences for codec data is impractical and unscalable. Instead, we construct the preference codec dataset by contrasting the golden codec tokens against synthetic codec tokens.
Concretely, we randomly sample $N$ speech-text pairs $P = \{(\mathbf{s},\mathbf{x})\}_{i=1}^{N}$ from LibriSpeech dataset,
where $\mathbf{s} = (s_1,...,s_{|s|})$ is the speech and $\mathbf{x} = (x_1,...,x_{|x|})$ is the corresponding transcript and  N is 50000.
For each speech $\mathbf{s}$, we adopt pretrained SpeechTokenizer to extract discrete representations and denote the tokens of first RVQ layer as golden AR tokens $\mathbf{y_g}$.
For the corresponding transcript $\mathbf{x}$, the AR model $p_{\theta_0}^{ar}$ takes it as input to generate synthetic AR tokens $\mathbf{y_s}$.
Following these steps, we can get the preference codec dataset $D_{pf} = \{(\mathbf{x},\mathbf{y_g},\mathbf{y_s})\}_{i=1}^{N}$.

\noindent\textbf{Human Verification}~
To validate the quality of constructed preference codec dataset, we perform human verification by randomly sampling 100 entries from $D_{pf}$ and employing the same procedure outlined in section~\ref{sec:pre:downstream} to convert $y_g$ and $y_s$ back into speech. This allows humans to compare them side by side and choose the better speech in terms of both speech quality and voice similarity.
From results in Table~\ref{tab:human}, we can conclude that humans prefer speech reconstructed from golden AR tokens over that from synthetic AR tokens, indicating that the constructed preference codec dataset effectively aligns with human preferences.

\begin{table}[t]
    \setlength{\abovecaptionskip}{0.3cm}
    % \setlength{\belowcaptionskip}{-0.2cm}
    % \renewcommand{\arraystretch}{1.2}
    % \Large
    \centering
    \resizebox{0.4\columnwidth}{!}{
    \begin{tabular}{c|c|c}
        \toprule
        Golden Win & Tie & Golden Lose \\
        \midrule
        $71\%$ & $21\%$ & $8\%$  \\
        \bottomrule
    \end{tabular}}
    \caption{Comparison between reconstructed speech from golden AR tokens versus synthetic AR tokens. 
    }
    \label{tab:human}
\vspace{-0.5em} 
\end{table}

\subsection{Preference Optimization}~
\label{sec:method:learn}
% \noindent We explore different preference-aware optimization strategies to align codec language models with preference dataset, including Chain-of-Hindsight~\cite{liu2023chain}, Direct Preference Optimization~\cite{rafailov2023direct}, RLHF-PPO~\cite{ouyang2022training} and Best-of-N Sampling.

In this section, we introduce how we conduct preference optimization to align codec language models using preference codec dataset, including Chain-of-Hindsight~\citep{liu2023chain}, Direct Preference Optimization~\citep{rafailov2023direct}, RLHF-PPO~\citep{ouyang2022training} and Best-of-N Sampling.

\noindent\textbf{Chain-of-Hindsight~(CoH)}~By converting various forms of feedback into sentences and integrating these with the respective responses,
CoH enables models to learn from both positive and negative feedback, allowing the identification and correction of negative attributes or errors.
At inference time, the model is guided to generate the desired outputs according to the feedback type in prompt.
In our case, we first convert feedback signals into a descriptive template and construct training data by combining responses with corresponding feedback template as follows:\\
$\mathbf{T_g}$ = "\textit{[Human]: Read this text and give me a high-quality speech response: $\{\mathbf{x}\}$ <eoh> [SpeechGPT]: $\{\mathbf{y_g}\}$ <eoa>.}" \\
$\mathbf{T_s}$ = "\textit{[Human]: Read this text and give me a low-quality speech response: $\{\mathbf{x}\}$ <eoh> [SpeechGPT]: $\{\mathbf{y_s}\}$ <eoa>.}" \\
The AR model is optimized via the negative log-likelihood loss on preference corpus $D_{pf}$ as follows:
\begin{align*}
    L_{COH} = -\mathbb{E}_{(x,y_g,y_s) \sim D_{pf}}[&\log p_{\theta_0}^{ar}(y_g|x, T_g) + \log p_{\theta_0}^{ar}(y_s|x, T_s) ]
\end{align*}
During inference phrase, we prompt the model with positive feedback in the form of ‘high-quality’ to guide the model in
generating favorable outputs.

\noindent\textbf{Direct Preference Optimization (DPO)}~Without using explicit
reward modeling or reinforcement learning, DPO can fine-tune the model to
align with human preferences. 
DPO considers the likelihood of preferred response over dispreffered response and optimizes the LLM model towards that objective.
The prompt template for DPO training is as follows:\\
$\mathbf{T}$ = "\textit{[Human]: Read this text and give me a speech response: $\{\mathbf{x}\}$ <eoh> [SpeechGPT]: $\{\mathbf{y}\}$ <eoa>.} " 
\\
In our case, the DPO loss can be formated as follows:
\begin{align*}
    L_{DPO}=- 
    \mathbb{E}_{(x,y_g,y_s) \sim D_{pf}} [ \log &\sigma ( \log \frac{p_{\theta}^{ar}(y_g | x, T)}{p_{ref}^{ar}(y_g | x, T)} - \log \frac{p_{\theta}^{ar}(y_s | x, T)}{p_{ref}^{ar}(y_s | x, T)} ) ]
\end{align*}
where $p_{ref}^{ar}$ is the reference model and initialize with $p_{\theta}^{ar}$.

\noindent\textbf{RLHF-PPO}~
RLHF methods involve training a reward model on a dataset reflecting human preferences. RL algorithms are then applied to adjust a language model's policy to favor responses that are highly rewarded, while ensuring minimal deviation from the original model's behavior.
With preference dataset $D_{pf}$, we can parameterize a reward model $r_{\phi}(x,y)$ and estimate the parameters via maximum likelihood. By treating the task as a binary classification, we utilize the negative log-likelihood loss:
\begin{align*}
    L_{rm} = \mathbb{E}_{(x,y_g,y_s) \sim D_{pf}}[\log \sigma(r_{\phi}(x,y_g)-r_{\phi}(x,y_s)]
\end{align*}
where $\sigma$ is the logistic function. The reward model $r_{\phi}(x,y)$ is initialized from AR model $p_{\theta}^{ar}$ with a linear layer atop the last Transformer layer to yield a single scalar prediction as the reward value.
During the RL stage,  we optimize the AR model against the reward model using PPO algorithm. Specially, we refine the AR model $p_{\theta_0}^{ar}$ as the following optimization problem:
\begin{align*}
    \max_{p_{\theta_0}^{ar}} ~&\mathbb{E}_{x \sim D_{pf},y \sim p_{\theta_0}^{ar}(y|x)}[r_{\phi}(x,y)] - \beta~\mathbb{D}_{kl}[~p_{\theta_0}^{ar}(y|x) || p_{ref}^{ar}(y|x)~]
\end{align*}
where $\beta$ represents a coefficient regulating the
extent of the KL penalty and $p_{ref}^{ar}$ is the reference model and initialize with $p_{\theta}^{ar}$.

\noindent\textbf{Best-of-N Sampling~(BoN)}~
With the reward model trained on the preference data, we implement a Best-of-N approach to enhance the quality of output codec tokens. Concretely, we sample $N$ responses
using the AR model. These responses are then evaluated by the reward model, and the one receiving the highest reward score is chosen as the final response to serve as input for NAR model.

\begin{algorithm}[t]
\caption{SpeechAlign}
\label{alg:speechalign}
\begin{algorithmic}
\STATE \textbf{Input:} $\{(s_i, x_i)\}_{i=0}^{N}$: Speech-Text Dataset, $m_{\phi}$: pretrained SpeechTokenizer model with parameter $\phi$, $p_{\theta_0}^{ar}$: AR model with parameter $\theta_0$, $T$: Number of iterations.
\FOR{$t = 0, \ldots, T-1$}
    \FOR{$i = 1, \ldots, N$}
        \STATE Generate golden AR token $y_{r_i} \sim m_{\phi}(\cdot|s_i)$
        \STATE Generate synthetic AR token $y_{s_i} \sim p_{\theta_t}^{ar}(\cdot|x_i)$
    \ENDFOR
    \STATE Update $\theta_{t+1}$ by performing preference-aware optimization on $\theta_{t}$ with $D_{pf_{i}} = \{(x_i,y_{r_i},y_{s_i})\}_{i=1}^{N}$
\ENDFOR
\STATE \textbf{Output:} $\theta_T$.
\end{algorithmic}
\end{algorithm}

\subsection{Iterative Self-Improvement}~
\label{sec:method:iterative}

Following the aforementioned steps results in an updated AR model, denoted as $p_{\theta_1}^{ar}$. Using this updated model, we can create a new preference codec dataset, $D_{pf}$. This dataset then serves as the basis for further improvement of the AR model through preference optimization.
The iterative self-improvement process of the AR model, as detailed in algorithm~\ref{alg:speechalign}, enables continuous calibration of the output distribution towards the authentic codec token distribution.

% As shown in algothrim~\ref{alg:speechalign}, the AR model can preform iteratively self-improvement and continuously calibrating the output distribution to authentic codec token distribution.

\section{Experiments}

\subsection{Setups}
\label{sec:041:setups}
\textbf{Data}~
For the continue finetuning stage in Section~\ref{sec:pre:background}, we use the LibriSpeech dataset. To construct the preference codec dataset, we randomly sample 50k speech-text pairs from LibriSpeech training set.
During the iterative process of SpeechAlign, we utilize the synthetic data generated in the most recent iteration and combine it with the newly produced synthetic data. As a result, the size of the synthetic dataset increases across iterations: starting at 50k in iteration 0, and expanding to 100k in iterations 1, 2, and 3. 

\noindent\textbf{Model}~
For the AR model, we further finetunes the pretrained SpeechGPT model in~\citep{zhan2024anygpt} on LibriSpeech dataset.
For the NAR model, we use the pretrained SoundStorm model in~\citep{zhan2024anygpt}.

\noindent\textbf{Training}~
For the continue finetuning stage in Section~\ref{sec:pre:background}, the batch size is set to 256, with a learning rate of 1e-5 and train for 3500 steps on 8 A100 80G GPUs.
For CoH finetuning, the batch size is set to 32, with a learning rate of 1e-5 and train for 12000 steps on 8 A100 80G GPUs.
For DPO finetuning, the batch size is set to 128, with a learning rate of 5e-7 and train for 2000 steps on 8 A100 80G GPUs.
For reward model training, the batch size is set to 32, with a learning rate of 1e-5 and train for 1000 steps on 8 A100 80G GPUs.
For PPO training, the batch size is set to 16, with a learning rate of 1e-5 and train for 1000 steps on 8 A100 80G GPUs.

\begin{table}[t]
\setlength{\abovecaptionskip}{0.3cm}
 \renewcommand{\arraystretch}{1.2}
 \Large
 \centering
 \resizebox{0.65\textwidth}{!}{
 \begin{tabular}{l|cc|cc}
 \toprule
 & \multicolumn{2}{c}{LibriSpeech} & \multicolumn{2}{c}{VCTK} \\
 Model & WER~($\downarrow$) & SIM~($\uparrow$) & WER~($\downarrow$) & SIM~($\uparrow$) \\
 \midrule
 Groundtruth & 4.0 & - & 1.7 & -   \\
 \midrule
 \multicolumn{5}{l}{\textit{Baselines}} \\
 SpeechAlign-sft & 7.2 & 0.87 &  8.8 & 0.79   \\
 Continue SFT & 8.0 & 0.88  & 9.8 & 0.80  \\
 \midrule
 SpeechAlign-CoH & 7.3& 0.89   & 10.2 & 0.81    \\
 SpeechAlign-RLHF-PPO & 7.1 & 0.89   & 8.5 & 0.80    \\
 SpeechAlign-BoN & 8.0 & 0.88   & 7.5 & 0.79   \\
 SpeechAlign-DPO-Iter1 & 6.7 & 0.88   &  8.5 & 0.82    \\
 SpeechAlign-DPO-Iter2 & 6.2 & 0.89  &  8.0 & 0.83    \\
 SpeechAlign-DPO-Iter3 & \textbf{6.0} & \textbf{0.90}  & \textbf{7.9} & \textbf{0.83}    \\
 
 \bottomrule
 \end{tabular}}
 \caption{Evaluation Results of zero-shot text-to-speech on LibriSpeech and VCTK. 
Each result is calculated as the average of ten separate evaluations.}
 \label{tab:main}
\end{table}

% \begin{table*}[ht]
% \setlength{\belowcaptionskip}{-0.2cm}
%  \renewcommand{\arraystretch}{1.2}
%  \Large
%  \centering
%  \resizebox{\textwidth}{!}{
%  \begin{tabular}{l|ccccc|ccccc}
%  \toprule
%  & \multicolumn{5}{c}{LibriSpeech} & \multicolumn{5}{c}{VCTK} \\
%  Model & WER~($\downarrow$) & SIM~($\uparrow$) & VISQOL~($\uparrow$) & QMOS~($\uparrow$)   & SMOS~($\uparrow$) & WER~($\downarrow$) & SIM~($\uparrow$)  & VISQOL~($\uparrow$) & QMOS~($\uparrow$) & SMOS~($\uparrow$) \\
%  \midrule
%  Groundtruth & 4.0 & -   & 1.7 & -    \\
%  \midrule
%  \multicolumn{11}{l}{\textit{Baselines}} \\
%  SpeechAlign-sft & 7.2 & 0.87   & 8.8 & 0.79    \\
%  Continue SFT & 8.0 & 0.88   & 9.8 & 0.80    \\
%  \midrule
%  Chain-of-Hindsight & 7.3& 0.89   & 10.2 & 0.81    \\
%  RLHF-PPO & 7.1 & 0.89   & 8.5 & 0.80    \\
%  Best-of-N & 8.0 & 0.88   & 7.5 & 0.79   \\
%  DPO~(Iteration 1) & 6.7 & 0.88   &  8.5 & 0.82    \\
%  DPO~(Iteration 2) & 6.2 & 0.89  &  8.0 & 0.83    \\
%  DPO~(Iteration 3) & \textbf{6.0} & \textbf{0.90}  & \textbf{7.9} & \textbf{0.83}    \\
 
%  \bottomrule
%  \end{tabular}}
%  \caption{Evaluation Results of zero-shot text-to-speech on LibriSpeech and VCTK. 
% Each result is calculated as the average of ten separate evaluations.}
%  \label{tab:main}
% \end{table*}

\subsection{Evaluation and Metrics}
\label{sec:042_metrics}
We conduct zero-shot TTS evaluation on LibriSpeech test-clean set and VCTK dataset.
For each speaker, we randomly selected a 3s utterance as the prompts while the textual content of a different utterance is used as the input text. 
To reduce the randomness in the evaluation process, we evaluate each model ten times and then calculate the average to obtain the final result.
The metrics we adopt are as follows:

\noindent\textbf{Word Error Rate (WER)} is utilized to assess the content accuracy of synthesized speech by calculating the distance between the synthesized speech’s transcription and the input text. We use Whisper medium-en model~\cite{radford2022robust} to transcribe the synthesized speech.

\noindent\textbf{Speaker Similarity (SIM)} evaluates the consistency of timbre between the synthesized and the prompt speech. This is measured by the similarity between the speaker embedding of generated speech and that of the speech prompt. The similarity calculation involves the following steps: 1) employing a speaker embedding extractor~\footnote{\url{https://huggingface.co/microsoft/wavlm-base-plus-sv}.}  to derive the speaker embeddings for both the generated and prompt speech, and 2) computing the cosine similarity between these normalized embeddings.

\noindent\textbf{Human Evaluation}~
We conduct comparative testing of various models' outputs against the baseline system's speech. During the evaluation phase, the evaluators are provided with prompt speech, the baseline system's speech, and our model's speech.
Human evaluators are tasked with determining which utterance sounded more natural and closer to the prompt speech. Evaluators have the option to choose either of the two utterances or indicate that they perceive them as equally natural. Each evaluation receives 6 ratings from 6 different human evaluators.

% \noindent\textbf{VISQOL}~measures the perceptual quality of speech. It uses a spectro-temporal measure of similarity between a reference and a test speech signal. The VISQOL scores range from 1 (the worst) to 5 (the best).~\citep{6309421}

% \noindent\textbf{Quality Mean Opinion Score (QMOS)} assesses the speech's quality and naturalness. For QMOS evaluations, we enlisted six native speakers as evaluators. The QMOS scale ranges from 1 to 5, where higher scores indicate superior speech quality.

% \noindent\textbf{Similarity Mean Opinion Score (SMOS)}~
% reflects the timbre similarity between speech prompts and
% generated speech. For the SMOS evaluations, six native speakers were recruited as evaluators. The SMOS scale ranges from 1 to 5, with higher scores indicating greater similarity in speech timbre.

\subsection{Main Results}
\label{sec:043_main}

\noindent\textbf{Preference Optimization Boosts Speech Generation}~
Figure~\ref{fig:comparsion} reveals that our preference-optimized models, SpeechAlign-BoN, SpeechAlign-RLHF-PPO and SpeechAlign-DPO-Iter1, significantly outperform the baseline model in win rates. 
As for objective evaluation, Table~\ref{tab:main} shows that the WERs of SpeechAlign-RLHF-PPO and SpeechAlign-DPO series models are lower than that of SpeechAlign-sft. This suggests that preference optimization can enhance the accuracy of content modeling. Furthermore, these models also achieved superior performance in Speaker Similarity, indicating that preference optimization can also improve the effectiveness of timbre modeling. 
These findings underscore the effectiveness of learning from human feedback in significantly improving the capabilities of codec language models across content, timbre, and audio quality dimensions.

\noindent\textbf{Speech Language Model Can Self-Improve Iteratively}~
The quantitative results in Table~\ref{tab:main} show that from Iteration 1, DPO contributes to enhancements in speech generation.
Iteration 2 further amplifies these improvements, with a notably significant impact on the WER.
And the trend of gradual enhancement is maintained in subsequent iterations.
By Iteration 3, there is a reduction in WER by 0.8 compared to the Baseline, and Speaker Similarity has increased to 0.9.
Figure~\ref{fig:comparsion} shows that SpeechAlign-DPO-Iter3 achieves a higher win rate compared to SpeechAlign-DPO-Iter1, indicating superior performance in qualitative evaluation.
This confirms that iterative DPO can consistently enhance the quality of the speech generated by the model. It demonstrates that SpeechAlign is an effective method for the speech language model to undergo continuous and efficient self-improvement.

\noindent\textbf{Genlization to Unseen Speakers}~
We also evaluate whether learning from human feedback would bring better speech generation when encountering unseen speaker in the preference data. 
We evaluate different models' performances on VCTK dataset. As shown in Table~\ref{tab:main}, SpeechAlign-RLHF-PPO, SpeechAlign-BoN, and SpeechAlign-DPO can still improves the generated speech across all metrics. We also observe similar improvements in subjective evaluation in Figure~\ref{fig:comparsion}. 
And iterative optimization can bring continuous improvement, suggesting that SpeechAlign can be generalized to unseen speakers.

\section{Analysis}

% \begin{figure*}[t] 
%     \centering 
%     \includegraphics[width=0.9\textwidth]{Figures/datasize.png} 
%     \caption{Performance of DPO with different data sizes.}
% \label{fig:dataset} 
% % \vspace{-0.5em} 
% \end{figure*}

\begin{figure*}[t] 
    \centering 
    \includegraphics[width=1\textwidth]{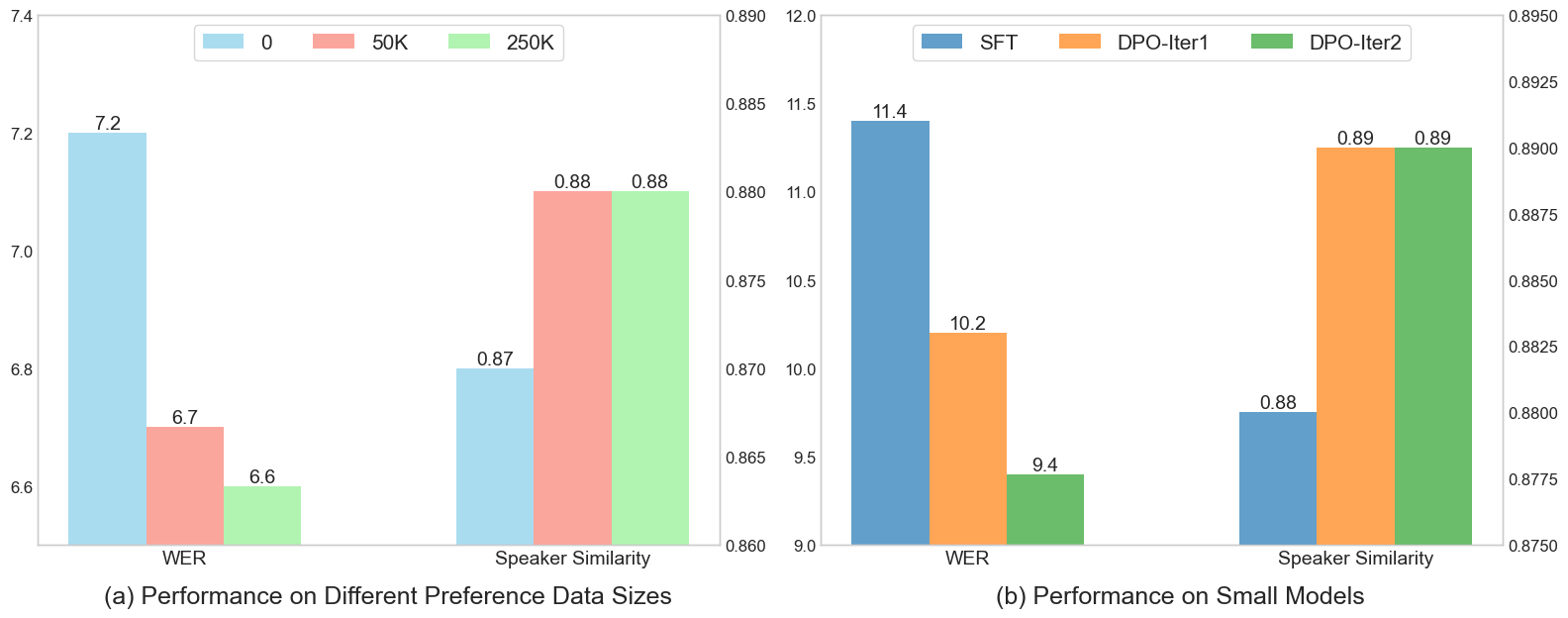} 
    \caption{\textbf{Left}: Performance of SpeechAlign across different preference data sizes. \textbf{Right}: Performance of SpeechAlign on small models.}
\label{fig:datasize_small} 
% \vspace{-0.5em} 
\end{figure*}

\subsection{Ablation Studies}
\label{sec:051:datasize}

\noindent\textbf{Preference Data Size}~
In this section, we examine the effect of different training data sizes on the performance of SpeechAlign.
We set the preference data size to be 0, 50k and 250k and generate the data accordingly.
We make sure that the larger dataset includes the smaller ones. After one epoch of DPO fine-tuning for each of these training sizes, we assess their performance in zero-shot TTS.
From Figure~\ref{fig:datasize_small} (a), we can observe notable improvement across all metrics with increasing training sizes from 0 to 50k, indicating that an increase in preference data can enhance the effectiveness of learning from feedback. However, employing 250k preference data through DPO does not result in larger performance gains, suggesting that there is a threshold beyond which additional data does not translate into better learning outcomes. This effect suggests that while increasing the size of preference data can initially lead to more effective feedback learning, there comes a point where the quality of data or the model's ability to utilize this data effectively becomes more critical than sheer volume.

\noindent\textbf{Comparsion With Continue SFT}~
The chosen samples in the preference dataset originate from ground truth AR tokens, therefore, during the preference optimization process, the model undergoes retraining on the data. To investigate whether the improvements from preference optimization or simply from continued training on ground truth data, we conduct experiments on continuing Supervised Fine-Tuning (SFT) from SpeechAlign-sft for comparison. 
We utilize the same ground truth AR tokens as those in the preference dataset, with training settings consistent with the DPO training described in Section~\ref{sec:041:setups}.
The findings presented in Table~\ref{tab:main} and Figure~\ref{fig:comparsion} reveal that continuing SFT does not enhance the quality of generated speech; rather, it may actually degrade performance. This indicates that the capability improvements gained from learning human feedback are attributable to preference-aware learning.

% \begin{figure*}[t] 
%     \centering 
%     \includegraphics[width=0.9\textwidth]{Figures/small.png} 
%     \caption{Performance of SpeechAlign on small model.}
% \label{fig:small} 
% % \vspace{-0.5em} 
% \end{figure*}

\subsection{SpeechAlign Works With Small Models}
\label{sec:052:small}
Current codec language models primarily rely on smaller AR models with fewer than 1 billion parameters~\citep{wang2023neural,borsos2022audiolm}.
So we investigate whether SpeechAlign can bring improvements for smaller AR models.
We utilize an AR model with 130 million parameters that features a 12-layer transformer decoder-only architecture with embedding dimension 768 and 16 heads.
We train the AR model on Multilingual LibriSpeech dataset for 500000 steps with batch size 1152 and learning rate 1e-5.
We adopt the NAR model in Section~\ref{sec:pre:background}.
For DPO training, the preference data size is 50k and we following the training setting in Section~\ref{sec:041:setups}.
We evaluate the zero-shot TTS performance on LibriSpeech test-clean dataset.
Figure~\ref{fig:datasize_small} (b) shows that in the first iteration, DPO decreases the WER from 11.4 to 9.3 and boosts Speaker Similarity from 0.87 to 0.88. This indicates that learning from human feedback can notably improve speech generation in smaller models.
As the process advances, it's evident that the effectiveness of DPO isn't just limited to initial gains. 
With subsequent iterations, we observe a consistent upward trend in WER though no more improvement in speaker similarity.

\subsection{Preference Optimization Bridges the Distribution Gap}
\label{sec:053_visual}
As depicted in Figure~\ref{fig:tnse} (b), after alignment, the representations of golden AR tokens and synthetic AR tokens merge into a single cluster without significant distribution differences. This demonstrates that the distribution gap can be bridged by learning from human feedback. Along with the results in Table~\ref{tab:main} and Figure~\ref{fig:comparsion}, we observe that as the distribution gap diminishes, the model's capability in speech generation improves. This proves the effectiveness of learning from human feedback in calibrating model's output distribution. By reducing the inconsistencies between the training and inference phases, the model can more accurately capture the features of the target distribution, resulting in more natural and accurate speech generation.

% \subsection{Comparison with Continue SFT}~

\section{Related Work}
\noindent\textbf{Neural Codec Language Models}~
AudioLM~\citep{borsos2022audiolm} is the pioneering model in introducing codec
codes for language modeling, adopting a hierarchical strategy that combines semantic and acoustic modeling. VALL-E~\citep{wang2023neural}, another innovative neural codec language model, is trained to produce discrete codes based on EnCodec~\citep{défossez2022high}, enabling the generation of high-quality, personalized speech from just a 3-second sample of an unseen speaker's speech.
AudioPalm~\citep{rubenstein2023audiopalm} and VioLA~\citep{wang2023viola} advance codec language modeling by merging text and audio tokens, facilitating the simultaneous processing and generation of text and speech. 
However, all existing codec language models have been trained through supervised methods. SpeechAlign is the first work enable codec language models to learn from human feedback.

\noindent\textbf{Learning From Human Feedback}~
Learning from human feedback recently became a critical step in the LLM training such as ChatGPT~\citep{openai2023gpt4} and LLaMA~\citep{touvron2023llama}. 
Existing human preference alignment
methods for LLM include RLHF~\citep{ouyang2022training,bai2022training}, contrastive learning~\citep{rafailov2023direct} and Chain-of-Hindsight~\citep{liu2023chain}.
Multi-modal large language models have also made remarkable performance~\citep{liu2023llava,zhu2023minigpt4,li2024groundinggptlanguage,zhang-etal-2023-speechgpt,zhan2024anygpt}. Improving multi-modal large language models 
with human feedback has received extensive attention. MusicRL~\citep{cideron2024musicrl} aligns music generation to human preferences by reinforcement learning, Baton~\citep{liao2024baton} improves audio generation by learning from human feedback and \citep{lee2023aligning} aligns text-to-image models using human feedback. \citep{liu2021reinforcement} proposes to improve emotional text-to-speech via reinforcement learning. However,
our work explores using human feedback to align codec language models for speech generation with human preference.

\noindent\textbf{Self-Improvement}~
SPIN~\cite{chen2024selfplay} enables the LLM to self-improve without additional human data or feedback from stronger
LLMs by generating its own training data from its previous iterations and refining its policy by discerning
these self-generated responses from those obtained from human-annotated data.
SPIN-Diffusion~\cite{yuan2024selfplay} makes the diffusion
model engage in competition with its earlier versions, facilitating an iterative self-improvement
process. 
~\citep{xu2024advancing} adopts RLHF to improve LLM translation quality by optimizing reward models by distinguishing
between human and machine translations.
Similarly, our work converts weak speech language models stronger learners through self-improvement iteratively.

\section{Conclusion}
This paper first analyzes the distribution gap existing in current neural codec language models and propose to solve it by learning from human feedback.
To avoid the need for additional human-annotated preference data, construct a
preference codec dataset contrasting golden codec tokens against synthetic tokens.
Then we conduct preference optimization to align codec language models to human preference. Subjective and objective evaluation results prove the effectiveness of SpeechAlign to continuously converting weak codec language models to stronger ones.

\section{Limitations and Future Works}
\noindent\textbf{Fine-grained Reward Signals from Real-World Human Preferences}~
Current preference datasets capture overall preferences, while speech preferences can be multi-faceted, including aspects like sound quality, rhythm, and timbre. Considering human preferences from these various dimensions can enhance speech generation capabilities in a more detailed manner. Additionally, gathering high-quality, real human preference data could be more effective than the current methods of preference dataset collection, as it allows for a nuanced understanding of user preferences that can lead to more targeted and efficient improvements in speech generation technologies.

\noindent\textbf{Preference Optimization of the NAR Models}~
In current practices, preference optimization is employed to enhance the capabilities of AR models. Nonetheless, the codec decoder also grapples with inconsistency problems during training and inference, stemming from distribution gaps between golden NAR tokens and synthetic NAR tokens. Therefore, applying preference optimization to calibrate the output distribution of NAR models is worth exploration.

\bibliography{custom}
% \bibliographystyle{acl_natbib}

% \bibliography{custom}
\bibliographystyle{neurips_2023}

% \appendix
% \section{Appendix}
% \input{Sections/090_appendix}

\end{document}